# CellLineNet: End-to-End Learning and Transfer Learning For Multiclass Epithelial Breast cell Line Classification via a Convolutional Neural Network

**Darlington A. Akogo [1], Vincent Appiah [1,2], Xavier-Lewis Palmer [1,3]**

1. minoHealth AI Labs, 00233, SCC Inside, Weija, Accra, Ghana
2. West African Center for Cell Biology and Infectious Pathogens (WACCBIP), University of Ghana
3. Biomedical Engineering Institute, Old Dominion University, Norfolk, VA 23529, United States of America

**Abstract**

Computer Vision for Analyzing and Classifying cells and tissues often require rigorous lab procedures and so automated Computer Vision solutions have been sought. Most work in such field usually requires Feature Extractions before the analysis of such features via Machine Learning and Machine Vision algorithms. We developed a Convolutional Neural Network that classifies 5 types of epithelial breast cell lines comprised of two human cancer lines, 2 normal immortalized lines, and 1 immortalized mouse line (MDA-MB-468, MCF7, 10A, 12A and HC11) without requiring feature extraction. The Multiclass Cell Line Classification Convolutional Neural Network extends our earlier work on a Binary Breast Cancer Cell Line Classification model. CellLineNet is 31-layer Convolutional Neural Network trained, validated and tested on a 3,252 image dataset of 5 types of Epithelial Breast cell Lines (MDA-MB-468, MCF7, 10A, 12A and HC11) in an end-to-end fashion. End-to-End Learning enables CellLineNet to identify and learn on its own, visual features and regularities most important to Breast Cancer Cell Line Classification from the dataset of images. Using Transfer Learning, the 28-layer MobileNet Convolutional Neural Network architecture with pre-trained ImageNet weights is extended and fine tuned to the Multiclass Epithelial Breast cell Line Classification problem. CellLineNet simply requires an imaged Cell Line as input and it outputs the type of breast epithelial cell line (MDA-MB-468, MCF7, 10A, 12A or HC11) as predicted probabilities for the 5 classes. CellLineNet scored a 96.67% Accuracy.

# Introduction

Breast cancer remains a leading diagnosed cancer for American women, with detected diagnoses in the hundreds of thousands and deaths in the tens of thousands, yearly as of 2017 (DeSantis et Al 2017). In facilities within countries that lack sophisticated means of detection, low cost, accessible tools of identification become ever the more important. We present an expansion of a tool that can be used with brightfield images of cancerous and non-cancerous cell lines, requiring no further preparation; this is expanding on our previous paper, "End-to-End Learning via a Convolutional Neural Network for Cancer Cell Line Classification", in which we offer a more complex model that distinguishes between multiple non-cancerous and cancerous cell lines, demonstrating the modularity as needed, per experiment, expanding the capability of automated identification and classification of known cancer cell lines to any lab with a brightfield microscope and camera (Akogo et al, 2018)

Convolutional Neural Network which were initially developed some 30 years ago (Fukushima, 1980, 1983, 1987) re-emerged in recent years mainly due to increase in data and computational power. Alex Krizhevsky and Ilya Sutskever work on the ImageNet Large Scale Visual Recognition Competition in 2012 (ILSVRC-2012) was very influential in the re-emergence of Convolutional Neural Networks. Since their re-emergence, they've been applied to a wide range of Computer Vision problems, from Object Detection to Image Segmentation problems (Liang-Chieh Chen et al., 2014, J. Redmon et al., 2015, S. Ren et al., 2015) and to specific domains like Medical Image Analysis (Shadi Albarqouni et al., 2016, Mark J. J. P. van Grinsven et al., 2016, Lin Yang et al., 2017, Andre Esteva et al., 2017).

Our model, CellLineNet is a 31-layer Convolutional Neural Network that's trained, validated and tested on a 3,252 image dataset of 5 types of Epithelial Breast cell Lines in order to be able to classify Breast Cancer Cell Lines of the types, MDA-MB-468, MCF7, 10A, 12A, and HC11. CellLineNet extends the 28-layer MobileNet Convolutional Neural Network architecture. It is also an expansion of our earlier 6-layer Convolutional Neural Network Binary Classifier trained, validated and tested on 1,241 images of MDA-MB-468 and MCF7 breast cancer cell line (Akogo DA and Palmer XL., 2018), which had a simpler architecture based on our ScaffoldNet architecture (Akogo et al., 2018).

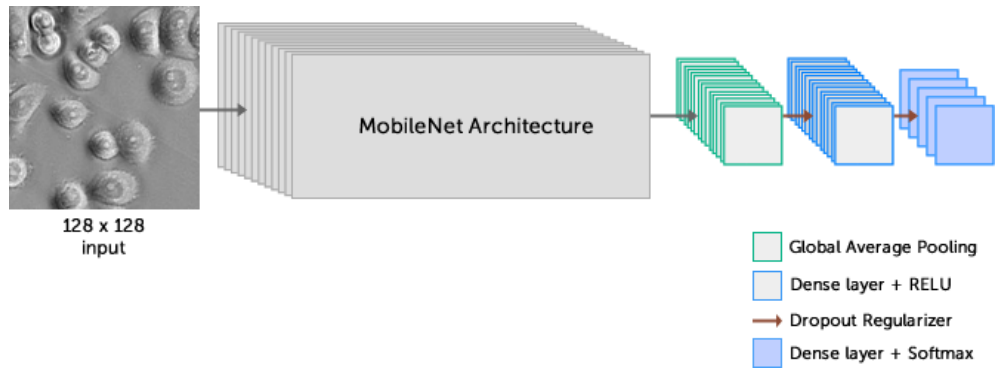

**Figure 1**. The architecture of CellLineNet. The input image is propagated through the MobileNet architecture then through a Global Average Pooling layer and Dense layer, with Dropout Regularizer applied in between. And then finally, 5 units Dense output layer preceded by another Dropout Regularizer.

## CellLineNet Architecture

We used a 31-layer Convolutional Neural Network that extends and fine tunes the 28-layer MobileNet Convolutional Neural Network architecture (Howard et al., 2017) without its first full convolutional layer (as shown in Figure 1). MobileNet is a light weight deep neural network meant to effectively maximize accuracy whilst still being mindful of constrained computing resources. The MobileNet architecture mainly uses Depthwise Separable Convolutions (as shown in Figure 2), which is a form of factorized convolutions which factorize a standard convolution into a depthwise convolution (SIfre et al., 2014) and a 1×1 convolution called a pointwise convolution. Depthwise Convolutions are used to apply a single filter per input channel, and Pointwise Convolution is then used to create a linear combination of the output of the depthwise layer. As shown in Figure 3, each layer is followed by Batchnorm and ReLU nonlinear function with the exception of the final fully connected layer.

We then added a Global Average Pooling layer to reduce the spatial dimensions of our tensor (Min Lin et al., 2013). Dropout Regularizer was then introduced with a fraction rate of 0.25 to further prevent overfitting (Srivastava et al., 2014). 512 unit densely-connected Neural Network layer was then added followed by another Dropout Regularizer with a fraction rate of 0.25. Finally, the 5 unit densely-connected Neural Network output layer was added. CellLineNet's final output layer uses a Softmax activation function:

$$\sigma(z)_j = \frac{e^{z_j}}{\sum_{k=1}^{K} e^{z_k}}$$

where $z$ is a vector of the inputs to the output layer (we have 6 output units, so there are 5 elements in $z$). And again, $j$ indexes the output units.

This squashes the raw class scores into normalized positive values, then outputs them as separate probabilities for each of our classes(MDA-MB-468, MCF7, 10A, 12A and HC11), where all the probabilities add up to 1.

Using Transfer Learning, the learned weights of MobileNet pre-trained on the ImageNet dataset (Russakovsky et al., 2014) were used for CellLineNet. Adam optimization algorithm was used with the standard parameters ($\beta1$ = 0:9 and $\beta2$ = 0:999) for training (Kingma and Ba, 2014). A learning rate ($\alpha$) of 0.001 and mini-batches of 32 were used. As training started on our 3,252 images Epithelial Breast cell Lines dataset, the weights were fine tuned in order for the whole architecture to adapt to Multiclass Breast Cancer Cell Line Classification task.

| Type / Stride | Filter Shape | Input Size |
|---|---|---|
| Conv / s2 | 3 × 3 × 3 × 32 | 224 × 224 × 3 |
| Conv dw / s1 | 3 × 3 × 32 dw | 112 × 112 × 32 |
| Conv / s1 | 1 × 1 × 32 × 64 | 112 × 112 × 32 |
| Conv dw / s2 | 3 × 3 × 64 dw | 112 × 112 × 64 |
| Conv / s1 | 1 × 1 × 64 × 128 | 56 × 56 × 64 |
| Conv dw / s1 | 3 × 3 × 128 dw | 56 × 56 × 128 |
| Conv / s1 | 1 × 1 × 128 × 128 | 56 × 56 × 128 |
| Conv dw / s2 | 3 × 3 × 128 dw | 56 × 56 × 128 |
| Conv / s1 | 1 × 1 × 128 × 256 | 28 × 28 × 128 |
| Conv dw / s1 | 3 × 3 × 256 dw | 28 × 28 × 256 |
| Conv / s1 | 1 × 1 × 256 × 256 | 28 × 28 × 256 |
| Conv dw / s2 | 3 × 3 × 256 dw | 28 × 28 × 256 |
| Conv / s1 | 1 × 1 × 256 × 512 | 14 × 14 × 256 |
| 5× Conv dw / s1 | 3 × 3 × 512 dw | 14 × 14 × 512 |
| 5× Conv / s1 | 1 × 1 × 512 × 512 | 14 × 14 × 512 |
| Conv dw / s2 | 3 × 3 × 512 dw | 14 × 14 × 512 |
| Conv / s1 | 1 × 1 × 512 × 1024 | 7 × 7 × 512 |
| Conv dw / s2 | 3 × 3 × 1024 dw | 7 × 7 × 1024 |
| Conv / s1 | 1 × 1 × 1024 × 1024 | 7 × 7 × 1024 |
| Avg Pool / s1 | Pool 7 × 7 | 7 × 7 × 1024 |
| FC / s1 | 1024 × 1000 | 1 × 1 × 1024 |
| Softmax / s1 | Classifier | 1 × 1 × 1000 |

**Figure 2**. Original architecture of MobileNet as shown in the MobileNet paper. The architecture is based on Depthwise Separable Convolutions.

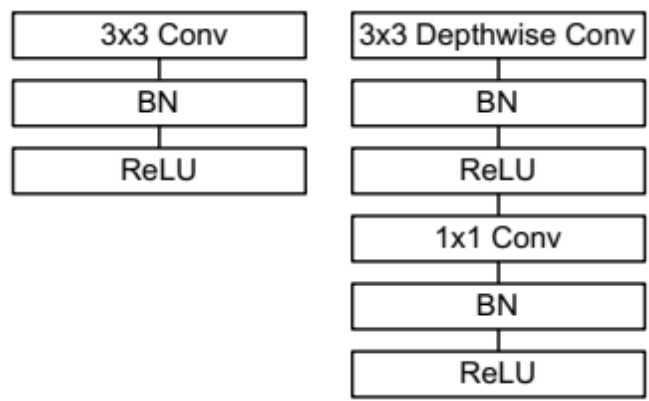

**Figure 3**. As shown in MobileNet paper. Left: Standard Convolution Layer followed by Batchnorm and ReLU nonlinear function. Right: Each Depthwise Convolution layer and Pointwise Convolution layer is followed by Batchnorm and ReLU nonlinear function.

| | | |
|---|---|---|
| 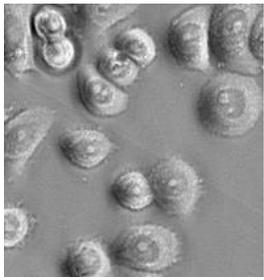 | 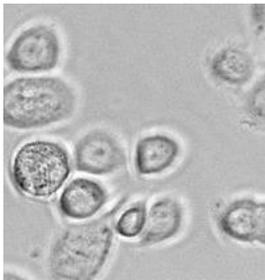 | MDA-MB-468 |
| 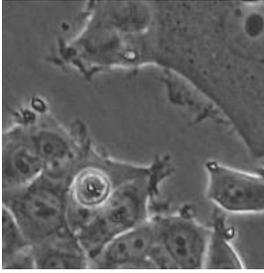 | 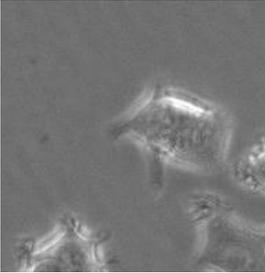 | MCF7 |
| 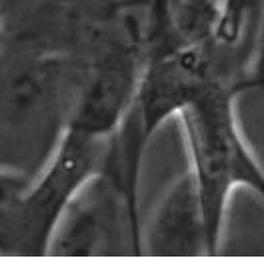 | 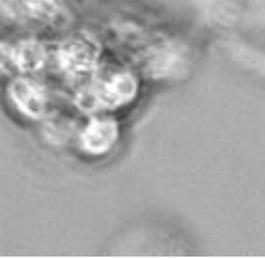 | 10A |
| 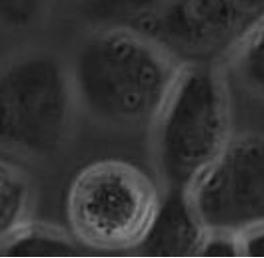 | 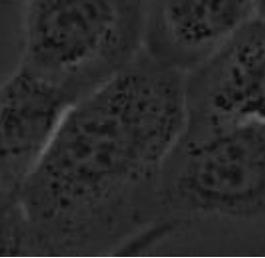 | 12A |
| 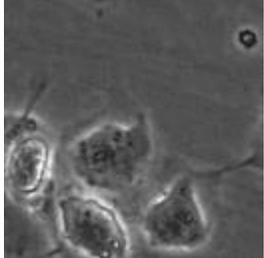 | 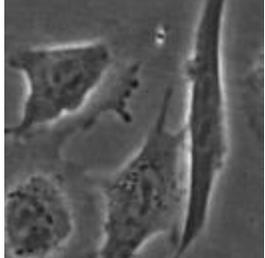 | HC11 |

**Figure 4.** Samples of the imaged Breast Cancer Cell Lines belonging to the 5 classes (MDA-MB-468, MCF7, 10A, 12A and HC11) used in training CellLineNet. As can been seen, there exist large variations of samples within each class and some similarities across the various classes.

## Data

Our dataset was a collection of 3,252 images of MDA-MB-468, MCF7, 10A, 12A and HC11 Epithelial Breast cell Lines collected under varied imaging conditions. This dataset was used in the training, validation, and testing of CellLineNet. The samples as seen in Figure 4 above vary within each class and yet share certain similarities across the various classes. Shape and clustering patterns give slight cues toward identification and classification, but could still pose difficulty for human eyes for work en masse, depending on confluency and other micro environmental conditions, lighting, and phase. The MCF7, MDA-MB-468, MCF-10A, MCF-12A, and HC11 cell lines are all epithelial cell lines, meaning that they line outer and inner surfaces of tissues; specifically they are mammary lines. The first four are human tissues, with the first two of those, the MCF7 and MDA-MB-468 cell lines being adenocarcinomas, cancerous cells that occur in glandular tissue. The other two, MCF-10A and MCF-12A, represent normal human epithelial tissues. Each are useful in a variety of studies on treating breast cancer. The last cell line, HC11, is found in mice. As shown in Figure 5 below, the dataset contains 664 MDA-MB-468 adenocarcinomic human breast cell images, 577 MCF7 adenocarcinomic human breast cell images, 684 HC11 normal mouse cell images, 703 10A normal human cell line images and 634 12A normal human cell line images. Each cell line was imaged from separate locations, with varied lighting, at 400X via brightfield microscopy from 6-well cell plates. Images were subsequently cut into 128x128 pixel images, which altogether reflected varying imaging conditions.

We split the 3,252 breast cancer cell line images dataset into *Training set*, *Validation set* and *Testing set* with a *8:1:1 ratio* (2,606, 323, 323), respectively. The images were all reshaped to 128 x 128 pixels and they were transformed by Standardization in order to achieve a similar range. All 3,252 epithelial breast cell line images were further augmented with random horizontal flips, width shifts, height shifts, 5° rotations, and zooms, in order to make our dataset contain more variation which ensures a more generalized trained and tested model.

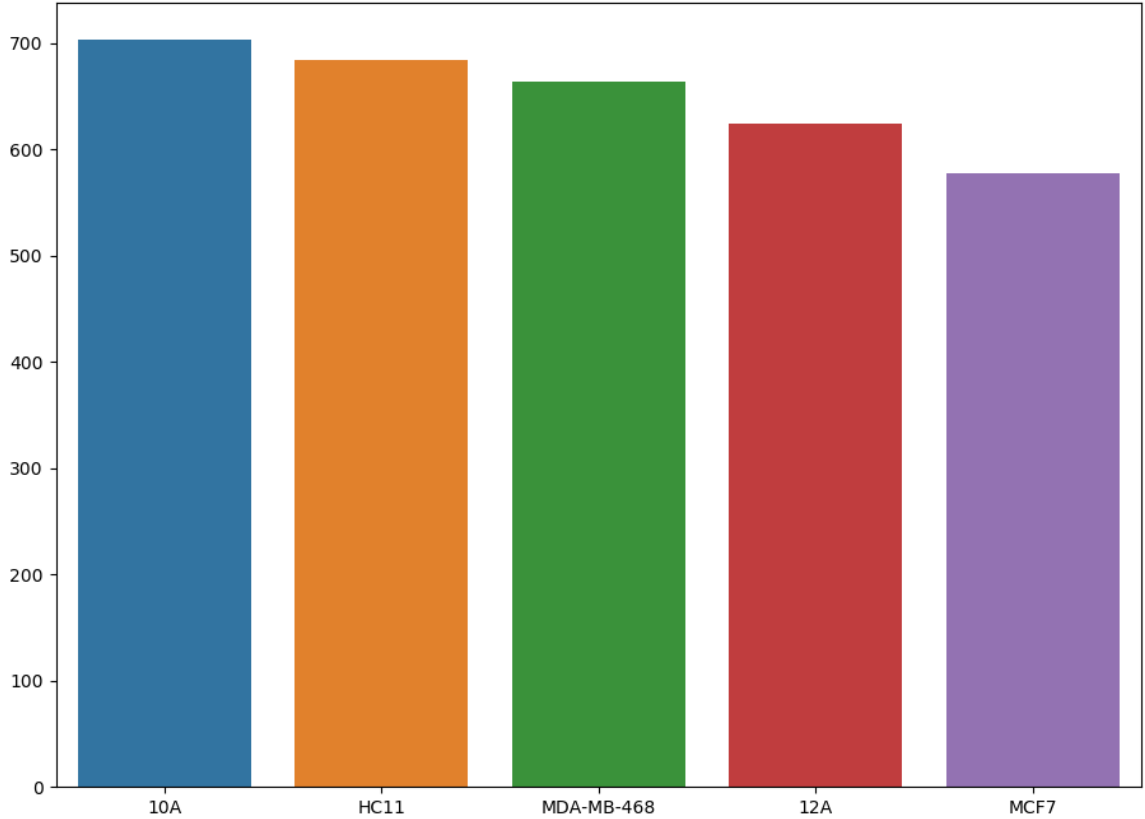

**Figure 5.** Sorted in descending order, the dataset contains 703 10A epithelial breast cell line images, 684 HC11 breast cancer cell line images, 664 MDA-MB-468 breast cancer cell line images, 634 12A breast cancer cell line images and 577 MCF7 breast cancer cell line images.

## Training and Validation

With our 2,606 epithelial breast cell line images Training set, CellLineNet was trained with Adam optimization algorithm. The loss function used was the *Cross-Entropy loss function.* The metric used during training was the *Accuracy Classification Score,* with the formula:

$$\text{accuracy}(y, \hat{y}) = \frac{1}{n_{\text{samples}}} \sum_{i=0}^{n_{\text{samples}}-1} 1(\hat{y}_i = y_i)$$

where the $\hat{y}_i$ is the predicted output for $i$-th sample $y_i$ is the (correct) target output computed over $n_{\text{samples}}$

With 3 epochs of training and hyperparameter tuning with 323 epithelial breast cell line images Validation set, CellLineNet's performance was improved. The first epoch led to a 95.56%

Accuracy Score on the Validation set and a 0.1586 Cross-Entropy loss as shown in Figure 6 below. This demonstrates the power of Transfer Learning and Fine Tuning.

And after an additional 2 epoch, CellLineNet's performance improved further which led to an Accuracy Score of 97.22% and Cross-Entropy loss of 0.1039. With Early Stopping set with a patience of 4 epochs, training haltered on the 3rd epoch. Even with Early Stopping adjusted or removed, training beyond 3 epochs within the next 10 or so epochs offered no improvement in loss.

|  | 1st Epoch | Final (3rd) Epoch |
|---|---|---|
| Accuracy Score | 95.56% | 97.22% |
| Cross-Entropy Loss | 0.1586 | 0.1039 |

**Figure 6**. Accuracy Score and Cross-Entropy loss of CellLineNet as recorded after the first epoch and final epoch (3rd epoch).

## Testing and Results

After Training and Validation with 3 epochs, CellLineNet true performance was evaluated with the 323 epithelial breast cell line images Test set.
CellLineNet's performance on the Test set was;
**Accuracy score: 96.67%**
**Cross-Entropy loss: 0.0749**

## Related Work

Earlier works that exist within the domain of Computer Vision for automated analysis of cells and tissues often segment cells from images and then extracts features like size and shape from such cells. Such features are then used to train Computer Vision systems that analyze these features using Machine Learning and other Machine Vision algorithms. Examples include a system where they features are extracted from segmented blood cells and then classified via Multilayer Perceptrons (Wei Lin et al., 1998). Other examples include localization of sub-cellular components via threshold adjacency statistics which are then analyzed by support vector machine (Hamilton NA et al., 2007). And grading of cervical intraepithelial neoplasia by extracting geometrical features that are analyzed using a combination of computerized digital image processing and Delaunay triangulation analysis (Keenan SJ, et al., 2000). Others compare extracted features and raw pixel densities analyzed via Bayesian Classifier, K-Nearest Neighbors, Support vector machine, and Random Forest (Timothy BL et al., 2016). CellLineNet however, is an End-to-Learning and Transfer Learning system which used learned general visual features from a large general Image Classification task and dataset, and then fine-tuned, identified and learned important visual features and regularities on its own pertaining Epithelial Breast cell Line Classification. This drastically simplifies the development of automated Computer Vision systems for cell and tissue analysis whilst maximizing their capabilities by allowing them discover important regularities on their own rather than being restricted to features we extract manually.

## Conclusion and Outlook

We developed CellLineNet, a 31-layer Convolutional Neural Network trained, validated and tested on a 3,252 image dataset of 5 types of Epithelial Breast cell Lines (MDA-MB-468, MCF7, 10A, 12A and HC11) in an end-to-end fashion. CellLineNet extends, transfers, and fine tunes the 2 8-layer MobileNet Convolutional Neural Network architecture with pre-trained ImageNet weights. Our Convolutional Neural Network is trained on a 2,606 Epithelial Breast cell Line image set, validated on a 323 Epithelial Breast cell Line image set and tested on a 323 Epithelial Breast cell image set. CellLineNet's final performance during evaluation was an Accuracy score of 96.67% and a Cross-Entropy loss of 0.0749. We believe that CellLineNet further extends the promise of automated systems for analysis and classification of other cancerous and normal cell lines of other diseases cases, which we may continue exploring more of in upcoming works. It also potentially can help lower barriers for care in less equipped labs.